\documentclass{ecai}
\usepackage{graphicx}
\usepackage{latexsym}
\usepackage{float}
\usepackage{amsmath}
\usepackage{xcolor}
\newtheorem{definition}{Definition}
\usepackage[bb=px]{mathalfa}

\begin{document}

\begin{frontmatter}

\title{CERM: Context-aware Literature-based Discovery via Sentiment Analysis}

\author[B]{\fnms{Julio Christian}~\snm{Young}\orcid{0000-0002-0483-370X}
\thanks{Corresponding Author. Email: julio.christian.young@gmail.com.}}
\author[A]{\fnms{Uchenna}~\snm{Akujuobi}\orcid{0000-0002-7102-7994}
}

\address[A]{Sony AI, Sony Research Inc.}
\address[B]{Graduate School of Advanced Science and Technology, Tokyo Denki University}

\begin{abstract}
Driven by the abundance of biomedical publications, we introduce a sentiment analysis task to understand food-health relationship. Prior attempts to incorporate health into recipe recommendation and analysis systems have primarily focused on ingredient nutritional components or utilized basic computational models trained on curated labeled data. Enhanced models that capture the inherent relationship between food ingredients and biomedical concepts can be more beneficial for food-related research, given the wealth of information in biomedical texts. Considering the costly data labeling process, these models should effectively utilize both labeled and unlabeled data. This paper introduces Entity Relationship Sentiment Analysis (ERSA), a new task that captures the sentiment of a text based on an entity pair. ERSA extends the widely studied Aspect Based Sentiment Analysis (ABSA) task. Specifically, our study concentrates on the ERSA task applied to biomedical texts, focusing on (entity-entity) pairs of biomedical and food concepts. ERSA poses a significant challenge compared to traditional sentiment analysis tasks, as sentence sentiment may not align with entity relationship sentiment. Additionally, we propose CERM, a semi-supervised architecture that combines different word embeddings to enhance the encoding of the ERSA task. Experimental results showcase the model's efficiency across diverse learning scenarios.

\end{abstract}

\end{frontmatter}

\section{Introduction}

\emph{We are what we eat - Ludwig Feuerbach}. This quote means that the food we eat can have either positive or negative effects on us. Hence, we should aim to consume food that brings us both health and happiness \cite{drees2022we}. In recent years, the interest in healthy lifestyles and healthy diets has surged, especially after the pandemic \cite{Coronavirus_exp,Poinski2020Consumer}.
Ideally, advice regarding an appropriate diet for an individual should be consulted with a nutritionist, considering their expert knowledge for planning a personalized and sustainable diet according to their needs. However, consulting with a nutritionist can often be time-consuming and require additional funds, making it less accessible. Previous studies have attempted to develop diet recommendation systems by training various artificial intelligence (AI) models on the massive food-related data available \cite{banerjee2019,bhat2021,iwendi2020,Kim2020,nigar2019,phanich2010}. While some studies focus on dietary recommendation systems targeting specific audiences (e.g., dietary planning for people with certain health conditions \cite{bhat2021,phanich2010} or age groups \cite{banerjee2019,nigar2019}), few studies target a broader audience \cite{iwendi2020,Kim2020}. 

In \cite{iwendi2020}, researchers developed a dietary recommendation system considering various factors such as demographic information, health conditions, and nutritional needs. The system accurately predicted meal suitability by representing it as a binary classification problem. However, due to its reliance on a neural network for prediction, the system lacked the ability to provide explanations for its decisions. Another study by Kim \& Chung \cite{Kim2020} combined symbolic AI and neural network approaches to produce a hybrid system that achieves the same goal but can explain its decisions. 
Due to the additional inference process by the symbolic AI approach, the resulting recommendations can be inherently explained. 
Although \cite{Kim2020} has successfully addressed the explainability issue in the recommendation system, there is still room for improvement to produce a better solution. 
Previous studies \cite{iwendi2020,Kim2020} have primarily focused on macronutrients and their impact on human health, neglecting the significance of micronutrients. However, the recommendation process should also consider essential factors like food-disease relationships, nutritional components, and ingredients, as outlined in \cite{maggini2018}. Neglecting these components may lead to unintended consequences, such as recommending allergenic meals to users or suggesting the consumption of micronutrients that are unsuitable for a user's medical condition.

A dietary recommendation system that can consider all the information in a meal concerning the user's medical condition can be realized based on two other supporting systems. The first system should be responsible for breaking down all the nutritional information in a meal. Then, based on such information, the second could infer the relationship between each element and the user's medical condition. Research on the first system has been extensively conducted, but research on the second is limited. In addition, research in this field often prioritizes the development of databases \cite{chauhan2020focus,MCKILLOP2021596,rothwell2016systematic,scalbert2011databases} that summarize existing relationships rather than adaptive systems that can learn and incorporate new information. Arguably, such a system is much needed given that research related to food science is still developing; thus, knowledge about the effects of compounds on diseases can change based on the latest studies. 

With abundant food science-related research articles, learning relationships between food ingredients and biomedical concepts can be structured as a literature-based sentiment analysis task. Specifically, this task involves learning the relationship between two entities, $e_{1}$ and $e_{2}$, given a short text (sentence) $s$ where the entities are mentioned. This problem is different from a regular sentiment analysis task since the sentiment of $s$ may not necessarily represent the sentiment of $e_{1}$ and $e_{2}$. For instance, consider a sentence \emph{``The daily consumption of Ginger is beneficial to the body but ineffective for Diarrhea.''}. Although the overall sentiment of the sentence is positive, the relationship between \emph{``Ginger''} and \emph{``Diarrhea''} have a contrasting sentiment. This example demonstrates that the overall sentiment of a sentence may not always indicate the sentiment of the relationship between entities mentioned within it. Furthermore, some sentences involve multiple entity pairs, further complicating the task. Hence, compared to regular sentiment analysis, this task presents a greater challenge. To address this issue, we propose modeling this problem as Entity Relationship Sentiment Analysis (ERSA).

\begin{definition}
    \it
    {\bf Entity Relationship Sentiment Analysis}: Given a sentence $s$ and two entities $e_{1}$ and $e_{2}$, where $e_{1} \neq e_{2}$, the goal is to determine the sentiment polarity of the relationship between $e_{1}$ and $e_{2}$ given $s$, where the relationship can be classified as either positive, negative, or neutral.
\end{definition}


In this study, we also propose CERM - a \textbf{C}ontext-aware \textbf{E}ntity \textbf{R}elationship Prediction \textbf{M}odel to address the ERSA problem more effectively. Inspired by promising results in previous studies \cite{alghanmi-etal-2020-combining,alharbi-lee-2021-multi,alharbi2021enhancing,d2020bert}, we combine the abilities of static and contextualized word embeddings models to generate richer representations of the problem inputs. While BERT, as a contextualized word embeddings model, is used to represent the features of the input sentence ($s$), a static word embeddings model is used to represent the features of each entity ($e_{1}$ and $e_{2}$). The combination of these two models is believed to produce richer representations because static word embeddings are better at capturing general semantic relationships between words (e.g., antonyms and synonyms), while contextualized embeddings are better at capturing more subtle semantic relationships (e.g., negation and sarcasm) \cite{alghanmi-etal-2020-combining,alharbi-lee-2021-multi,alharbi2021enhancing,d2020bert}. Furthermore, the two models also produce different types of information, and their combination can result in a more comprehensive representation. 
Furthermore, considering the significant resources required for data labeling, we propose a semi-supervised learning (SSL) strategy with the proposed model to leverage unlabeled data during the learning process.

The contributions of this paper include:
\begin{itemize}
    \item We introduce a new problem called Entity Relationship Sentiment Analysis (ERSA), which enables concept pair relationship inference from a wealth of food science-related research articles through sentiment analysis.
    \item We propose Context-aware Entity Relationship Prediction Model (CERM) for the ERSA task. Our experiments demonstrate that CERM, our proposed model, outperforms the state-of-the-art semi-supervised text classification methods with application to the ERSA task. We have also showcased the effectiveness of CERM by evaluating its performance on the Aspect-based Sentiment Analysis (ABSA) task.
    \item We introduce our dataset to facilitate further research in this literature-based sentiment analysis domain.
\end{itemize}

\section{Related Works}

\subsection{Fine-grained Sentiment Analysis Tasks}

Fine-grained sentiment analysis is a task in natural language processing that goes beyond the basic classification of sentiment and seeks to identify more nuanced sentiment levels in text. Entity-sentiment analysis (ESA) \cite{mitchell-etal-2013-esa}, aspect-based sentiment analysis (ABSA) \cite{saeidi-2016}, and multi-entity sentiment analysis (ME-ABSA) \cite{Yang_Yang_Wang_Xie_2018} are some examples of fine-grained sentiment analysis tasks. ESA involves identifying the sentiment associated with specific entities mentioned in a text, such as people, organizations, or products. ESA is more complex than regular sentiment analysis due to the presence of multiple entities in a text, and the sentiment towards one entity may contradict the sentiment towards another. For instance, based on a person's tweet, the sentiment towards a specific player may be positive, but the sentiment towards the management may be negative \cite{mitchell-etal-2013-esa}.

On the other hand, the ABSA task focuses on assigning multiple sentiments to a given text based on a list of aspects. For example, in \cite{saeidi-2016}, researchers define an ABSA task to detect multiple aspects (price, safety, transit, and general) in a comment toward a tourist destination and predict sentiment values for each aspect. As an ABSA task requires a more fine-grained understanding of the language used in the text, it is also considered a more challenging task than a regular sentiment analysis task. In \cite{Yang_Yang_Wang_Xie_2018}, researchers extend the ABSA task to define the ME-ABSA task, where the focus is on identifying the sentiment with specific aspects of multiple entities mentioned in a text. As the ME-ABSA task deals with multiple entities and aspects, it is considered more challenging than ESA and ABSA tasks. 

Similar to the previously mentioned tasks, the ERSA task can be classified as a fine-grained sentiment analysis task. While ERSA focuses on predicting the sentiment of the relationship between two named entities, whether explicitly or implicitly expressed in a sentence, ME-ABSA  deals with the sentiment of multiple entities and a predefined set of aspects. The critical distinction is that ERSA does not have a predefined set of relationships between entities, making it challenging to map or encode as a ME-ABSA task. In ERSA, the primary objective is to predict the sentiment of the entity relationship itself rather than the sentiment of an entity with respect to a specific aspect, which is the primary goal of ME-ABSA. Therefore, ERSA cannot be directly mapped or encoded as an ME-ABSA task.

\subsection{Semi-supervised Learning on Text Classification}

Semi-supervised learning is motivated by the time-consuming and expensive process needed for data labeling, especially for data that requires domain-specific knowledge. Furthermore, a much larger pool of unlabeled data is often available than labeled data. There have been several applications of SSL in text classification. In \cite{lee2013pseudo}, researchers proposed a mechanism for training a machine learning model that utilizes both labeled and unlabeled data. The method involves initially training the model on the labeled data and then using it to generate pseudo-labels for the unlabeled data. The key insight behind this approach is that the model's predictions on the unlabeled data will likely be accurate for at least some examples.

In \cite{xie2020uda}, researchers defined an SSL algorithm by incorporating data augmentation techniques on the unlabeled data. By constraining the model's predictions for the original and augmented versions of the unlabeled data, the model can learn to be consistent. For example, in a traditional sentiment analysis task, if the model predicts a sentiment value of a sentence like \emph{``Scientific research shows that ginger could help ease a sore throat''} to be `positive,' the model can further enhance its learning by applying a consistency loss that encourages consistent predictions for its augmented versions. The proposed method was also shown to perform well against other SSL algorithms for various text classification tasks. The researchers in \cite{sohn-fm} further extended the idea by combining consistency regularization and pseudo-labeling techniques. Similarly to \cite{xie2020uda}, FixMatch uses consistency learning to encourage the model to make similar predictions for slightly perturbed versions of the same input. Instead of using all predictions on the unlabeled data, the FixMatch method only considers pseudo-labels with high confidence during the model training.

In our proposed method, we incorporate consistency regularization on the unlabeled data, following the promising results of the previous two studies. Instead of using BackTranslation (BT) to generate noisy examples, we utilize Easy Data Augmentation (EDA) in our proposed method as it better preserves the original text's meaning and requires lower computational resources. Moreover, considering the input from the problem domain, we also employ cosine embedding loss to generate a better model to learn the relevance of keywords and their corresponding text.



\section{ERSA Dataset}
The ERSA dataset is a dataset for entity relationship sentiment analysis. This dataset is extracted from the publication text of papers in the PubMed dataset \footnote{https://pubmed.ncbi.nlm.nih.gov/download/}.

From the abstracts and full paper text of Pubmed publications, we extract 
sentences with one or more mentions of predefined entities. These predefined entities which are obtained from multiple sources\footnote{\url{http://ctdbase.org/downloads/#alldiseases}}\footnote{\url{https://www.nlm.nih.gov/mesh/meshhome.html}}\footnote{\url{https://omim.org/}}\footnote{\url{https://www.ncbi.nlm.nih.gov/geo/info/download.html}}\footnote{\url{https://pubchem.ncbi.nlm.nih.gov/docs/downloads}} can be placed in 6 groups: Genes, Disease, Chemical compounds, Nutrition, and Food ingredients. We then generate entity pairs from entities that appear in the same sentence. Thereby obtaining a dataset that focuses on the relationship between two entities, $e_{1}$ and $e_{2}$, in a given sentence, $s$. 
Due to the high cost of data labeling, we only select $50,000$ entity pairs with their corresponding sentences for labeling. We randomly selected entity pairs with corresponding sentences for labeling to achieve a representative distribution of entity pairs in the dataset. We then invite external data curators to assign the sentiment of selected entity pairs given their corresponding sentences. This labeling is done using the Amazon Mechanical Turk System\footnote{https://www.mturk.com
}. Each entity pair and respective sentences are labeled by 3 different people as either positive, negative, or neutral. When there are disagreements, we use the majority label, and if there is no majority consensus, the data point is removed. After cleaning and post-processing the labeled data, we obtained $11,366$ labeled entity pairs. The final labeled dataset has $2,890$ positive, $3,191$ negative, and $3,011$ neutral entity pairs given their corresponding sentences. See Table \ref{table1} for data statistics.

Given the labeled data, we randomly divided it into a 70/30 train-test split, where 70\% of the data was used for training, and the remaining 30\% was reserved for testing. Additionally, we augmented the training set with all available unlabeled data. Table \ref{table1} presents descriptive statistics for the dataset.
\begin{table}
\begin{center}
{\caption{The statistics of ERSA dataset}\label{table1}}
\begin{tabular}{|l | c | c|} 
 \hline
 \textbf{Property names} & \textbf{Training Set} & \textbf{Testing Set} \\ [0.5ex] 
 \hline
 Labelled data count & 9092 & 2274 \\
  \enspace - Positive  & 2890 & 696  \\
  \enspace - Negative  & 3191 & 785  \\
  \enspace - Neutral  & 3011 & 793  \\
 \hline
 unlabeled data count & 50000 & - \\
 \hline
 Word count & & \\
  \enspace - Average  & 35.59 &  35.20 \\
  \enspace - Min  & 3 & 3  \\
  \enspace - Max  & 1008 & 603  \\
 \hline
 \# of unique entities & 18184 & 4548  \\
  \enspace - Chemicals  & 7823 & 1971  \\
  \enspace - Consumables  & 2450 & 657  \\
  \enspace - Diseases  & 7467 & 1820  \\
  \enspace - Nutrients  & 115 & 28  \\
  \enspace - Gene  & 329 & 72  \\
 \hline
\end{tabular}
\end{center}
\end{table}
Based on the statistics, it can be seen that the data has a relatively balanced label distribution. Also, it can be observed from the data that each data point contains a unique entity that does not repeat in other data points. This condition can lead to several issues (e.g., limited context and inconsistent quality) in the training process of the static word embeddings model if the learning process is only conducted on labeled data. Given that for the labeled data, each entity is only present in a single sentence; there may be insufficient contextual information available to extract relevant features (a limited context issue). Additionally, the lack of diversity in the quality and relevance of the sentences for each entity may significantly impact the model's performance (inconsistent quality). Hence the need for a semi-supervised approach in learning.

\section{The Proposed Method}

\begin{figure*}
\centerline{\includegraphics[height=2.8in]{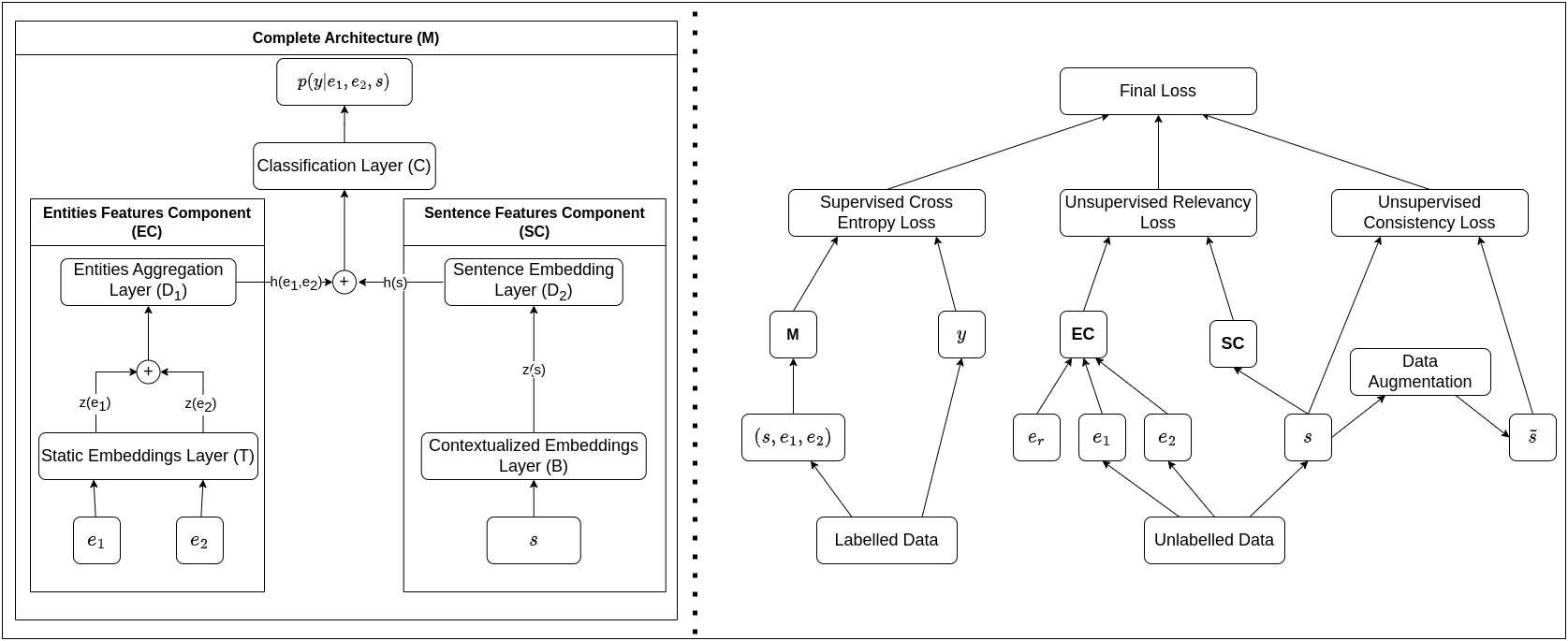}}
\caption{The proposed CERM model. The left side shows the proposed model's architecture ($M$) with its underlying entities features component ($EC$) and sentence features component ($SC$) for ERSA task. The right side shows the semi-supervised learning approach} \label{architecture}
\end{figure*}

\subsection{The Architecture Overview}
The proposed method operates based on a combination of static and contextualized word embeddings. Figure \ref{architecture} illustrates the model architecture of the proposed method.
\newline Given an input of entity pair $\{e_{1}, e_{2}\}$ and corresponding sentence, $s$, CERM extracts features for $e_{1}$ and $e_{2}$ using static word embeddings layer, $T$, and for $s$ using contextualized word embeddings layer, $B$.
The entity combination layer $D_{1}$ merge the individual embedding $z(e_{1})$ and $z(e_{2})$ of the entities in the entity pair to generate a rich representation of the entity pair.
Similarly, the sentence embedding layer $D_{2}$ takes input from $B$ to generate the sentence context embedding.
The layers $D_{1}$ and $D_{2}$ are modeled as MLP layers and 
are intentionally designed to have the same dimensions to facilitate the learning process on unlabeled data proposed in this research. For a more stable model, we apply the ReLU activation to the output of the linear layers.

In the architecture, $T$ and $B$ can take the form of any pre-trained static word embedding model and contextualized word embeddings, respectively. $\oplus$ is a concatenation operation. For our experiments, we chose to use a FastText model \cite{bojanowski-2017} that was trained on both labeled and unlabeled training data for $T$ and BioBERT, a pre-trained BERT model that was trained on a large biomedical text corpus \cite{lee2020biobert}, for $B$. We chose the FastText model for its ability to handle out-of-vocabulary (OOV) words through subword information and BioBERT due to its suitability for the numerous biomedical terms and language used in the dataset. Finally, the classification layer $C$ concatenates the learned entity pair embeddings and the sentence embeddings from $D_{1}$ and $D_{2}$ respectively, and projects it into class probabilities using $\mathrm{softmax}$.


\subsection{Task Specific Fine-Tuning for Contextualized Embeddings}
\label{ce-fine-tuning}
Taking inspiration from previous studies in \cite{taesun2020}, we apply a similar transformation to the input sentence $s$ prior to feature extraction 
to enhance the overall model's learning performance. The transformation is carried out by introducing a new token to the input of the contextualized embedding layer $B$. 
This strategy aims to introduce an additional signal that emphasizes the importance of the selected part to the model \cite{taesun2020}. Thereby enabling the model to concentrate on critical elements, thus enhancing its performance. The proposed method transforms the input sentence $s$ by including a special token, `[CTX].' Specifically, we add this token before and after each keyword's appearance, $e_{1}$ and $e_{2}$, in a sentence $s$. 

Furthermore, fine-tuning of the $B$ model is performed solely on the last few attention layers, as suggested by previous studies \cite{houlsby2019,vucetic2022}. This strategy aims to optimize resource usage and minimize the time required for fine-tuning while preventing overfitting. This is achieved by allowing the $B$ model to preserve its pre-trained knowledge. We apply the gradient accumulation strategy \cite{zhuang-2021} for the parameter update during training. The gradient accumulation strategy smooths out the learning process by reducing the impact of noisy updates and allows the model to generalize on GPUs with limited memory.

\subsection{Semi-Supervised Sentiment Analysis Task}

In this subsection, we describe how the model can perform semi-supervised learning for the ERSA task. We denote the set of input as $X = \{X_l, X_u\}$, where $X_l$ is the set of labeled data samples and $X_u$ is the set of unlabeled samples. The respective set of labeled and unlabeled samples are composed of individual data point $x$, which represents a tuple $(e_{1}, e_{2}, s)$, where $e_{1}$ and $e_{2}$ are two entities mentioned in a sentence $s$. The labeled data $X_l$ has a corresponding set of labels $Y$ such that a labeled data tuple is given as $(x, y); \quad x \in X_l, y \in Y $. We define the classification layer as $f(., \theta)$, with output prediction of $p_{\theta}(y|x)$, which predicts $y$ based on the input $x$, where $\theta$ represents the model parameters. The target is to infer the sentiments of unseen data by learning from labeled and unlabeled data.

\subsubsection{Supervised Sentiment learning loss}
\label{sup_los_sec}

Given the labeled data samples $x \in X_l$ and the corresponding labels $y^{*} \in Y$, the prediction probability scores $p_{\theta}(y|x)$ are obtained from the output of the classification layer $f(., \theta)$. The cross-entropy (CE) loss is computed based on the predicted labels defined below.

\begin{equation}
\mathcal{L}_{CE}(X,Y,\theta)=-\frac{1}{N}\sum_{i=1}^{N}\sum_{j=1}^{L}y^{*}_{ij}\log p_{\theta}(y_{ij}=1|x_{i})
\label{celoss}
\end{equation}

\noindent where $N$ is the number of labeled samples, $L$ is the number of classes, $y^{*}_{ij}$ is the ground truth label for the $i$-th sample and $j$-th class (either 0 or 1), and $p_{\theta}(y_{ij}=1|x_{i})$ is the predicted probability of the $i$-th sample belonging to the $j$-th class, given the input $x_i$ and the model parameters $\theta$. The outer sum computes the average over all samples in $X$. The model will learn a parameter $\theta$ that minimizes $CE$ loss.

\subsubsection{Unsupervised Consistency Learning}
\label{con_loss_sec}

Inspired by a promising result of previous work in \cite{xie2020uda}, our proposed model also uses a data augmentation technique to enable learning from a set of unlabeled data, $X_u$. We denoted $\epsilon(x)$ as a function representing all augmented versions of $x$. We minimize the divergence metric between the two distributions as defined in equation \ref{conloss}.

\begin{equation}
\mathcal{L}_{con}(X_u,\theta)= \frac{1}{|X_u|} \sum_{x \in X_u} \mathbb{E}_{\tilde{x}\sim \epsilon(x)}[KL(p_{\theta}(y|x), p_{\theta}(y|\tilde{x}))]
\label{conloss}
\end{equation}

\noindent where $\tilde{x}$ is a perturbed version of $x$ sampled from the noise distribution $\epsilon$. The Kullback-Leibler (KL) divergence measures the difference between the predicted probability distribution of $p_{\theta}(y|x)$ and the predicted probability distribution of the perturbed version $p_{\theta}(y|\tilde{x})$. We encourage the model to produce similar outputs over the inputs by minimizing the consistency loss. However, unlike previous research that used TF-IDF word replacement and Back-Translation, our model uses Easy Data Augmentation (EDA) \cite{wei-zou-2019-eda} as function $\epsilon(x)$. Compared to TF-IDF word replacement, EDA generates more diverse augmented data while computationally less expensive than BT.


\subsubsection{Relevancy learning via Similarity Learning}
\label{cos_loss_sec}

In this subsection, we use the notation $h_{(e_{1},e_{2})}$ to represent the output features from $D_{1}$ corresponding to keywords $e_{1}$ and $e_{2}$, and $h_{s}$ to represent the output features from $D_{2}$ corresponding to sentence $s$. In addition to learning from the unlabeled data, $X_u$, via consistency loss, the model utilized cosine embedding loss for similarity learning. During the training process, the model learns the relevance between $\{e_{1}, e_{2}\}$, and $s$ by minimizing the cosine embedding loss between $h_{(e_{1}, e_{2})}$ and $h_{s}$, while also maximizing the loss between $h_{(e_{1}, e_{r})}$ and $h_{s}$, where $e_{r}$ is a randomly selected entity and $e_{r} \neq e_{1} \neq e_{2}$. Our similarity loss is defined as:

\begin{equation}
\mathcal{L}_{cos}(X_u,\theta)= \frac{1}{|X_u|} \sum_{(e_{1},e_{2},s) \in X_u} \mathbb{E}_{e_{r}\sim \Omega{(e)}}[\phi(e_{1},e_{2},e_{r},s)]
\label{cosloss}
\end{equation}

\noindent where $e_{r}$ is randomly selected from all keywords that appear in the dataset, $\Omega(e)$, and $\phi(e_{1}, e_{2}, e_{r}, s)$ is a cosine embedding loss defined in Equation  \ref{celdef}

\begin{align}
\phi(e_{1},e_{2},e_{r},s) = &1 - cos(h_{(e_{1},e_{2})},h_{s}) + \nonumber\\
&max(0, cos(h_{(e_{1},e_{r})},h_{s})- m).
\label{celdef}
\end{align}

\noindent In the Equation \ref{celdef}, $cos$ is a cosine similarity function to measure the similarity or dissimilarity between two input vectors, and $m$ is the margin that controls the degree of separation between the positive and negative pairs in the embedding space. We set the $m$ value to zero in our proposed model. The objective of the $\phi$ is to ensure that the model encodes the relationship between subject entities and the sentence meaningfully rather than simply memorizing patterns from each feature.

\subsubsection{Parameter Learning}

To train the network, we jointly optimize $\mathcal{L} = \mathcal{L}_{C}  + \mathcal{L}_{con} + \mathcal{L}_{cos}$, over the model parameters. Loss $\mathcal{L}_{C}$ is the classification loss as described in section \ref{sup_los_sec}, $\mathcal{L}_{con}$ is the consistency loss as described in section \ref{con_loss_sec}, and $\mathcal{L}_{cos}$ is the similarity loss as described in section \ref{cos_loss_sec}.


\section{Experiment}

To demonstrate the advantage of the proposed method for the ERSA task, we compared its classification performance against various supervised and semi-supervised learning methods. We evaluate the model performance using accuracy and f1 score on the testing set.

For the fully supervised methods, FastText or BERT was used to extract features from the text before feeding them into the classifier. To emphasize the encoding of the ERSA task, a special token `[CTX]' was inserted before and after each occurrence of the entities' keywords in the text before the feature extraction phase. 
We also evaluate the performance of the proposed model with and without fine-tuning the BERT for the task. The details regarding the feature extraction model used can be found in section \ref{featureextraction}, while the training settings of each classifier can be found in section \ref{classifierparams}.

In this paper, we evaluate the performance of the proposed model on four supervised methods:  logistic regression (LR), MLP, LSTM, and Bi-LSTM. 
To compare the performance of our proposed method against other SSL methods, we employed pre-trained BERT as a classifier and utilized Unsupervised Data Augmentation (UDA) \cite{xie2020uda}, Auxiliary Deep Generative Model (ADGM) \cite{maaloe-adgm}, and FixMatch \cite{sohn-fm} as the benchmarks.


\subsection{Feature Extractor Settings}
\label{featureextraction} Rather than utilizing an existing FastText model, we opted to train our model for the experiment, using a combination of labeled and unlabeled data from the training set and considering the specific language used for the task. There were $17,955$ sentences in total used in the training process. We had previously trained a FastText model on more sentences ($57,955$), but we found that the model trained on a smaller dataset achieved better performance, so we decided to train the model using a smaller number. The model is a skip-gram model, trained with negative sampling ($n=5$) and sub-word information ($n\_gram=[1-6]$) to produce an embedding with a size of $200$. The initial learning rate and context window are set to $0.1$ and $30$, respectively.

On the other hand, the BERT model used in the study is a pre-trained model trained on a large biomedical text corpus, BioBERT, which is available in \cite{lee2020biobert}. The utilization of BioBERT was conducted through two different settings. Firstly, by utilizing the model solely for feature extraction. 
And secondly, by jointly fine-tuning the BERT layer for this task. In the later subsection, we will refer to the BioBERT model that acts as a feature extractor as BERT and BERT-F for BioBERT with fine-tuning.

\subsection{Learning Parameters for the Classifier}
\label{classifierparams} 
The training process for each classification method involved experimentation with several parameters that could impact the model's performance. To accomplish this, we conducted an exhaustive grid search to find the optimal parameter settings. The best model was selected based on its performance on the test data using the f1-score macro average. PyTorch was used to implement all models except for the logistic regression (LR)         model. The scikit-learn implementation was used for the LR. For all models, the learning process utilized Adam optimizer with default parameters. For the baseline models, we implemented a hyper-parameter search and presented the results of the best configuration. For more information about the parameters for each model, please refer to section \ref{reproduce}.

\begin{table*}
\centering
{\caption{Comparison of results for the ERSA task. * denotes the semi-supervised methods using the pretrained BERT layer \textbf{only} as a feature extraction layer (i.e., the BERT model is not fine-tuned during training.) \label{table2}}}
\begin{tabular}{|l|r|r|r|r|r|}
\hline
          & \multicolumn{1}{l|}{F1-Score (Negative)} & \multicolumn{1}{l|}{F1-Score (Neutral)} & \multicolumn{1}{l|}{F1-Score (Positive)} & \multicolumn{1}{l|}{Macro F1} & \multicolumn{1}{l|}{Accuracy} \\ \hline
Logistic Regression & 0.67                                     & 0.64                                    & 0.67                                     & 0.66                          & 0.66                          \\
MLP                 & 0.68                                     & 0.65                                    & 0.68                                     & 0.67                          & 0.67                          \\
LSTM                & 0.66                                     & 0.63                                    & 0.64                                     & 0.64                          & 0.64                          \\
Bi-LSTM             & 0.66                                     & 0.64                                    & 0.65                                     & 0.65                          & 0.65                          \\ \hline
ADGM*               & 0.63                                     & 0.59                                    & 0.60                                     & 0.61                          & 0.61                          \\
UDA*                & 0.68                                     & 0.64                                    & 0.65                                     & 0.66                          & 0.66                          \\
FixMatch*           & 0.65                                     & 0.63                                    & 0.63                                     & 0.64                          & 0.64                          \\
CERM*    & 0.69                                     & 0.67                                    & 0.68                                     & 0.68                          & 0.68                          \\ \hline
UDA                 & \textbf{0.72}                            & 0.67                                    & \textbf{0.71}                            & 0.70                          & 0.70                          \\
FixMatch            & \textbf{0.72}                            & 0.68                                    & \textbf{0.71}                            & 0.70                          & 0.70                          \\
CERM     & \textbf{0.72}                            & \textbf{0.69}                           & \textbf{0.71}                            & \textbf{0.71}                 & \textbf{0.71}                 \\ \hline

\end{tabular}
\end{table*}

\subsubsection{Experiment Reproducibility}
\label{reproduce}

\textbf{Logistic Regression. }
All parameters were set to default except for the inverse of the regularization coefficient ($c$) and maximum iteration ($max\_iter$). The $max\_iter$ was set to 1000 while $c$ to 0.1. \newline
\textbf{MLP. }
The MLP model was designed with one hidden layer, where the number of neurons in the hidden layer was 200. ReLU activation function was used in the model and was trained using a dropout layer with a dropout rate of 0.5. The model was trained with a batch size of 32 with a learning rate of 5e-05. The maximum number of epochs was set to 50. \newline
\textbf{LSTM and Bi-LSTM. }
Both models were designed to have one hidden layer, with the learning rate, batch size, and the number of neurons in the hidden layer set to 1e-05, 64, and 200, respectively. The Bi-LSTM model concatenated the last hidden layer states from both forward and backward phases for classification, whereas the regular LSTM used only the forward phase.\newline
\textbf{Auxillary Deep Generative Model. }
For the ADGM implementation, the experiment was only conducted using BERT as the feature extractor. The encoder layer consists of three layers, with [512, 256, 128] hidden neurons. The number of auxiliary variables was set to match the number of latent variables (the third layer). The decoder layer mirrored the encoder layer with respect to the number of hidden neurons. During the training process, the maximum epoch was set to 400, and the batch size was set to 256. \newline
\textbf{BERT Fine-tuning with UDA/ FixMatch. }
The fine-tuning process was carried out using the parameters obtained from the proposed model to compare the model's performance under the same configuration. Besides that, the lambda value for unsupervised learning was set to one to calculate the final loss, and the threshold value of 0.5 was used for the FixMatch implementation. The epoch was set to 2.\newline
\textbf{The Proposed Model. }
In the proposed model, the batch size value was set to 64 with a learning rate and epoch of 2e-05. A new token, `[CTX],' was added to the tokenizer and used to encapsulate keywords in the text input for the BERT model. The number of neurons of $D_{1}$ and $D_{2}$ layers are set to 100. The data augmentation on the unlabeled data was performed using EDA by randomly removing, swapping, inserting, or replacing 20\% of the input text.\newline

\subsection{Result and Analysis}
Table \ref{table2} presents the results of all methods in the ERSA task.
The MLP model achieved the highest performance in the fully supervised learning settings, followed by the LR model. In such settings, the LSTM and Bi-LSTM models produced lower performance compared to the other two models. This result indicates that the BERT model used as a feature extractor already captures a significant amount of information necessary for the task, so using a more complex classification method only leads to the overfitting of the complete model.

The experiments on semi-supervised learning methods using BERT as just a feature extractor (denoted by '*') also showed similar results. As the ADGM implementation used an extra encoder-decoder and another complex network to capture auxiliary information on top of BERT, the trained model had an inferior performance. Furthermore, training the model using FixMatch or UDA also yields worse model performance, suggesting the inappropriateness of both methods for such settings. While the other two SSL methods failed to improve the classification model performance, CERM showed promising results by achieving a slightly better macro F1 score compared to the best result achieved in fully supervised settings. This result demonstrates how the extra complexity in our proposed model can effectively learn from the unlabeled data. Unlike the ADGM implementation, which suffered decreased performance due to the addition of extra complexity to the model, CERM showed the opposite result. Intuitively, adding an extra network on top of two embeddings that learn different representations can improve performance. This addition allows the model to learn richer representations tailored for the ERSA task through the combined embeddings.

Lastly, we also compared the performance of other semi-supervised learning methods with fine-tuned BERT model with our proposed method. Based on the results, it can be observed that all three semi-supervised learning methods outperformed the supervised methods with fine-tuning. The proposed model achieved the most substantial performance improvement, followed by the UDA and FixMatch methods. Our model's better ability to classify the neutral class was the main reason for its substantial performance, as shown by further analysis of each class F1-score. In this paper, we use 10k unlabeled samples. Adding more unlabeled data for all three SSL strategies resulted in worse model performance.
We postulate that for this dataset, due to the limited number of classes, a randomly sampled 10k unlabeled dataset is enough to generalize across the unlabeled data. Thus, adding more unlabeled data may hurt the model's performance, as the model may start to overfit.

\subsection{Model Prediction on Special Cases}

To demonstrate the model's performance on some non-trivial cases, we show models' predictions on sampled sentences in Table \ref{table3}. 
\newline In Table \ref{table3}, the CERM column represents the predictions made by our proposed model on the given text, while FM and UDA indicate the predictions made by Figure  fine-tuned BERT models using the FixMatch and UDA SSL strategies, respectively. The first two examples describe multiple entity-to-entity relationships, the next two have a misalignment on the overall sentiment w.r.t the entity-to-entity relationship sentiment, and the last sample expresses mixed sentiments on the same entity-entity relationship. For the last example, a robust model is expected to give a neutral prediction due to the sentiment conflict. Based on these examples, the proposed model accurately predicts entity relationship sentiment in the first four cases. In contrast, models with FM and UDA struggle in the third example, where sentence sentiment doesn't align with entity relationships. UDA model also faces difficulty with multiple entities in the sentence (example 1). However, we also notice the model sometimes fails to predict the correct label for neutral sentiments on sentences expressing mixed sentiment, similar to the other two models (example 5). 
We postulate that for such cases, incorporating the entity domains/types would enhance the prediction by infusing into the model an additional understanding of the entities.
\begin{table}
\begin{center}
{\caption{Model Performance on Special Cases. }\label{table3}}\begin{tabular}{|l | p{3cm} | l | l | l | l|} 
 \hline
 No. & Sentence & Actual & CERM & UDA & FM \\
 \hline
 1 & while \textbf{fenugreek} raise \textbf{testosterone}, there is no significant increase for participants that used tongkat ali and tribulus & + & \textcolor{blue}{+} & o & +\\
 \hline
 2. & consuming high-potassium foods like \textbf{honeydew} and banana excessively can result in \textbf{hyperkalemia}& - & \textcolor{blue}{-} & -& -\\
 \hline
 3. & there is still a lack of research on this area but a few suggest a good correlation between \textbf{garlic} and \textbf{sleep quality} & + & \textcolor{blue}{+} & o & o\\
 \hline
 4. & research on highly processed foods consumption is good for increase society awareness on how \textbf{refined sugar} can lead to \textbf{obesity} & - & \textcolor{blue}{-} & - & -\\
\hline
 5. & Study in B showed a significant increase in \textbf{testosterone} levels with \textbf{tongkat ali} consumption, while study in A reported a contrasting finding, suggesting a potential negative effect & o & \textcolor{red}{+} & + & -\\
 \hline 
\end{tabular}
\end{center}
\end{table}
\subsection{Performance on Another Dataset: ABSA Task}

To further validate the effectiveness of our method, we applied it to the task of Targeted Aspect-Based Sentiment Analysis for sentiment prediction, using the Sentihood dataset from a previous study \cite{saeidi-2016}. The dataset consists of $5,215$ annotated sentences that contain one or two mentions of location entities. The names of location entities in the dataset are masked with ``location1'' and ``location2'' throughout, so the task does not involve identifying and segmenting named entities. There are 12 aspects included in the dataset, but only the four most frequent aspects were utilized in the experiment. Likewise, to compare our method's sentiment prediction performance with existing methods, we will only consider these four aspects.

However, we conducted two testing scenarios since the provided dataset contains no unlabeled data while our proposed method is trained using a semi-supervised technique. The first scenario only utilized the architecture in the proposed method and conducted fully-supervised learning. Meanwhile, half of the training data was treated as unlabeled in the second scenario, and the semi-supervised learning process was applied. In both scenarios, the best model parameters were determined using the provided validation set, and the final performance of the model was evaluated on the test set using the accuracy metric, consistent with the methodology employed in the previous study. 

To encode sentiment prediction in the proposed method, we treat the entity, aspect, and sentence as $e_{1}$, $e_{2}$, and $s$. As the aspect is often not present in $s$, to match our problem setup, we concatenated the aspect $e_{2}$ at the end of $s$. We then add the `[CTX]' token before and after each occurrence of $e_{1}$ and $e_{2}$ in $s$. The static embedding, $T$,  model used in this task is the pre-trained FastText model on Common Crawl and Wikipedia data \footnote{Available at: https://fasttext.cc/docs/en/crawl-vectors.html}. Meanwhile, the contextualized embedding, $B$, model used is the pre-trained BERT on BookCorpus and Wikipedia \footnote{Available at: https://huggingface.co/bert-base-cased}. The number of neurons of $D_{1}$ and $D_{2}$ layers are set to 100. The best training parameters consisted of Adam optimizer with a batch size of 16, a learning rate of 5e-05, and an epoch of 8. To tackle the imbalanced class distribution problem in the dataset, we utilized the weighted loss method to update the weights. Table \ref{table4} compares our method and previous baseline methods.

\begin{table}
\begin{center}
{\caption{Comparison of results for the sentiment prediction in ABSA task. }\label{table4}}
\begin{tabular}{|l | c |} 
 \hline
 \textbf{Methods} & \textbf{Accuracy} \\ [0.5ex] 
 \hline
  LR-Left-Right** & 0.847  \\
  LR-Mask(ngram)**  & 0.853 \\
  LR-Mask(ngram+POS)** &  \textbf{0.875}  \\
  LSTM-Final** & 0.820  \\
  LSTM-Location** & 0.819 \\
 \hline
  The proposed method & \\
  \enspace - Semi-supervised & \textbf{0.874} \\
  \enspace - Fully-supervised & \textbf{0.885}\\
 \hline
 \multicolumn{2}{p{0.9\linewidth}}{\footnotesize **Retrieved from previous study in \cite{saeidi-2016}}
\end{tabular}
\end{center}
\end{table}

Based on Table 3, it can be observed that our fully-supervised trained model outperforms the best SOTA classification method in previous studies. Furthermore, in the semi-supervised learning setting, the model can achieve a relatively similar accuracy compared to the best classification method in the previous study by only utilizing half of the labeled data. These results demonstrate the benefits of combining pretrained static and contextualized embeddings in the proposed architecture and the advantages of using the proposed loss function, which allows the model to learn relevant information from unlabeled data.

\section{Conclusion}
In this paper, we propose a new sentiment analysis task called Entity Relationship Sentiment Analysis (ERSA) that focuses on capturing the sentiment between two entities. This task presents a more significant challenge than traditional sentiment analysis tasks, as the sentiment expressed in the sentence may not necessarily reflect the sentiment of the relationship between the entities. Furthermore, we propose an architecture combining two types of word embeddings to encode the ERSA task better. In addition to that, we also propose a method to enable the model to learn in a semi-supervised manner, considering the significant resources required for data labeling. The experimental results demonstrate that our proposed method consistently outperforms other semi-supervised learning methods across various learning scenarios. 

In our future work, we aim to enhance the relevancy learning loss by incorporating Multi-class N-pair loss \cite{NIPS2016_6b180037}. 
By doing so, we can emphasize the importance of the entity pairs in the sentence and encourage the model to increase the distance when incorrect words are chosen. By integrating Multi-class N-pair loss, we anticipate improving the model's ability to capture the relationships between entity pairs and ultimately enhance the overall relevancy learning process. In addition, we aim to explore various pre-trained transformer-based models to enhance the generalization of static and contextualized embedding combinations. It would be interesting to evaluate the performance of the proposed method by incorporating a distilled transformer-based model. This evaluation would provide insights into how the learning strategy employed in the proposed method operates on smaller models.

\bibliography{ecai}

\begin{thebibliography}{10}

\bibitem{alghanmi-etal-2020-combining}
Israa Alghanmi, Luis Espinosa~Anke, and Steven Schockaert, `Combining {BERT}
  with static word embeddings for categorizing social media', in {\em
  Proceedings of the Sixth Workshop on Noisy User-generated Text (W-NUT 2020)},
  pp. 28--33, Online, (November 2020). Association for Computational
  Linguistics.

\bibitem{alharbi-lee-2021-multi}
Abdullah~I. Alharbi and Mark Lee, `Multi-task learning using a combination of
  contextualised and static word embeddings for {A}rabic sarcasm detection and
  sentiment analysis', in {\em Proceedings of the Sixth Arabic Natural Language
  Processing Workshop}, pp. 318--322, Kyiv, Ukraine (Virtual), (April 2021).
  Association for Computational Linguistics.

\bibitem{alharbi2021enhancing}
Abdullah~I Alharbi, Phillip Smith, and Mark Lee, `Enhancing contextualised
  language models with static character and word embeddings for emotional
  intensity and sentiment strength detection in arabic tweets', {\em Procedia
  Computer Science}, {\bf 189},  258--265, (2021).

\bibitem{banerjee2019}
Anonnya Banerjee and Nishargo Nigar, `Nourishment recommendation framework for
  children using machine learning and matching algorithm', in {\em 2019
  International Conference on Computer Communication and Informatics (ICCCI)},
  pp. 1--6, (2019).

\bibitem{bhat2021}
Salliah~Shafi Bhat and Gufran~Ahmad Ansari, `Predictions of diabetes and diet
  recommendation system for diabetic patients using machine learning
  techniques', in {\em 2021 2nd International Conference for Emerging
  Technology (INCET)}, pp. 1--5, (2021).

\bibitem{bojanowski-2017}
Piotr Bojanowski, Edouard Grave, Armand Joulin, and Tomas Mikolov, `Enriching
  word vectors with subword information', {\em Transactions of the Association
  for Computational Linguistics}, {\bf 5},  135--146, (2017).

\bibitem{chauhan2020focus}
Shweta~Singh Chauhan, Deepak~Kumar Sachan, and Ramakrishnan Parthasarathi,
  `Focus-db: An online comprehensive database on food additive safety', {\em
  Journal of Chemical Information and Modeling}, {\bf 61}(1),  202--210,
  (2020).

\bibitem{drees2022we}
Betty~M Drees and Brandon Barthel, `We are what we eat', {\em Missouri
  Medicine}, {\bf 119}(5),  479, (2022).

\bibitem{d2020bert}
Ashwin~Geet d'Sa, Irina Illina, and Dominique Fohr, `Bert and fasttext
  embeddings for automatic detection of toxic speech', in {\em 2020
  International Multi-Conference on:“Organization of Knowledge and Advanced
  Technologies”(OCTA)}, pp. 1--5. IEEE, (2020).

\bibitem{Coronavirus_exp}
FoodCircle.
\newblock Coronavirus {Pandemic} {Triggers} {Surge} in {Demand} for {Healthy}
  {Foods}.
\newblock
  https://www.foodcircle.com/magazine/increase-healthy-foods-health-conscious-consumers.

\bibitem{houlsby2019}
Neil Houlsby, Andrei Giurgiu, Stanislaw Jastrzebski, Bruna Morrone, Quentin
  De~Laroussilhe, Andrea Gesmundo, Mona Attariyan, and Sylvain Gelly,
  `Parameter-efficient transfer learning for nlp', in {\em International
  Conference on Machine Learning}, pp. 2790--2799. PMLR, (2019).

\bibitem{iwendi2020}
Celestine Iwendi, Suleman Khan, Joseph~Henry Anajemba, Ali~Kashif Bashir, and
  Fazal Noor, `Realizing an efficient iomt-assisted patient diet recommendation
  system through machine learning model', {\em IEEE Access}, {\bf 8},
  28462--28474, (2020).

\bibitem{Kim2020}
Joo-Chang Kim and Kyungyong Chung, `Knowledge-based hybrid decision model using
  neural network for nutrition management', {\em Information Technology and
  Management}, {\bf 21}(1),  29--39, (Mar 2020).

\bibitem{lee2013pseudo}
Dong-Hyun Lee, `Pseudo-label: The simple and efficient semi-supervised learning
  method for deep neural networks', in {\em Workshop on challenges in
  representation learning, ICML 2013}, (2013).

\bibitem{lee2020biobert}
Jinhyuk Lee, Wonjin Yoon, Sungdong Kim, Donghyeon Kim, Sunkyu Kim, Chan~Ho So,
  and Jaewoo Kang, `Biobert: a pre-trained biomedical language representation
  model for biomedical text mining', {\em Bioinformatics}, {\bf 36}(4),
  1234--1240, (2020).

\bibitem{maaloe-adgm}
Lars Maaløe, Casper~Kaae Sønderby, Søren~Kaae Sønderby, and Ole Winther,
  `Auxiliary deep generative models', in {\em Proceedings of The 33rd
  International Conference on Machine Learning}, eds., Maria~Florina Balcan and
  Kilian~Q. Weinberger, volume~48 of {\em Proceedings of Machine Learning
  Research}, pp. 1445--1453, New York, New York, USA, (20--22 Jun 2016). PMLR.

\bibitem{maggini2018}
Silvia Maggini, Adeline Pierre, and Philip~C. Calder, `Immune function and
  micronutrient requirements change over the life course', {\em Nutrients},
  {\bf 10}(10), (2018).

\bibitem{MCKILLOP2021596}
Kyle McKillop, James Harnly, Pamela Pehrsson, Naomi Fukagawa, and John Finley,
  `Fooddata central, usda’s updated approach to food composition data
  systems', {\em Current Developments in Nutrition}, {\bf 5},  596, (2021).

\bibitem{mitchell-etal-2013-esa}
Margaret Mitchell, Jacqui Aguilar, Theresa Wilson, and Benjamin Van~Durme,
  `Open domain targeted sentiment', in {\em Proceedings of the 2013 Conference
  on Empirical Methods in Natural Language Processing}, pp. 1643--1654,
  Seattle, Washington, USA, (October 2013). Association for Computational
  Linguistics.

\bibitem{nigar2019}
Nishargo Nigar and Mohammed nazim uddin, `An intelligent children healthcare
  system in the context of internet of things', in {\em 2nd International
  Conference on Innovative Research in Science, Technology \& Management}, (06
  2018).

\bibitem{phanich2010}
Maiyaporn Phanich, Phathrajarin Pholkul, and Suphakant Phimoltares, `Food
  recommendation system using clustering analysis for diabetic patients', in
  {\em 2010 International Conference on Information Science and Applications},
  pp. 1--8, (2010).

\bibitem{Poinski2020Consumer}
M.~Poinski.
\newblock Consumer trends shifting toward health and wellness, {ADM} finds.
\newblock
  https://www.fooddive.com/news/consumer-trends-shifting-toward-health-and-wellness-adm-finds/584388/,
  sep 3 2020.

\bibitem{rothwell2016systematic}
Joseph~A Rothwell, Mireia Urpi-Sarda, Maria Boto-Ordo{\~n}ez, Rafael Llorach,
  Andreu Farran-Codina, Dinesh~Kumar Barupal, Vanessa Neveu, Claudine Manach,
  Cristina Andres-Lacueva, and Augustin Scalbert, `Systematic analysis of the
  polyphenol metabolome using the phenol-explorer database', {\em Molecular
  nutrition \& food research}, {\bf 60}(1),  203--211, (2016).

\bibitem{saeidi-2016}
Marzieh Saeidi, Guillaume Bouchard, Maria Liakata, and Sebastian Riedel,
  `{S}enti{H}ood: Targeted aspect based sentiment analysis dataset for urban
  neighbourhoods', in {\em Proceedings of {COLING} 2016, the 26th International
  Conference on Computational Linguistics: Technical Papers}, pp. 1546--1556,
  Osaka, Japan, (December 2016). The COLING 2016 Organizing Committee.

\bibitem{scalbert2011databases}
Augustin Scalbert, Cristina Andres-Lacueva, Masanori Arita, Paul Kroon,
  Claudine Manach, Mireia Urpi-Sarda, and David Wishart, `Databases on food
  phytochemicals and their health-promoting effects', {\em Journal of
  agricultural and food chemistry}, {\bf 59}(9),  4331--4348, (2011).

\bibitem{NIPS2016_6b180037}
Kihyuk Sohn, `Improved deep metric learning with multi-class n-pair loss
  objective', in {\em Advances in Neural Information Processing Systems}, eds.,
  D.~Lee, M.~Sugiyama, U.~Luxburg, I.~Guyon, and R.~Garnett, volume~29. Curran
  Associates, Inc., (2016).

\bibitem{sohn-fm}
Kihyuk Sohn, David Berthelot, Chun-Liang Li, Zizhao Zhang, Nicholas Carlini,
  Ekin~D. Cubuk, Alex Kurakin, Han Zhang, and Colin Raffel, `Fixmatch:
  Simplifying semi-supervised learning with consistency and confidence', in
  {\em Proceedings of the 34th International Conference on Neural Information
  Processing Systems}, NIPS'20, Red Hook, NY, USA, (2020). Curran Associates
  Inc.

\bibitem{vucetic2022}
Danilo Vucetic, Mohammadreza Tayaranian, Maryam Ziaeefard, James~J. Clark,
  Brett~H. Meyer, and Warren~J. Gross, `Efficient fine-tuning of bert models on
  the edge', in {\em 2022 IEEE International Symposium on Circuits and Systems
  (ISCAS)}, pp. 1838--1842, (2022).

\bibitem{wei-zou-2019-eda}
Jason Wei and Kai Zou, `{EDA}: Easy data augmentation techniques for boosting
  performance on text classification tasks', in {\em Proceedings of the 2019
  Conference on Empirical Methods in Natural Language Processing and the 9th
  International Joint Conference on Natural Language Processing
  (EMNLP-IJCNLP)}, pp. 6382--6388, Hong Kong, China, (November 2019).
  Association for Computational Linguistics.

\bibitem{taesun2020}
Taesun Whang, Dongyub Lee, Chanhee Lee, Kisu Yang, Dongsuk Oh, and Heuiseok
  Lim, `An effective domain adaptive post-training method for bert in response
  selection', {\em Proceedings of the Annual Conference of the International
  Speech Communication Association, INTERSPEECH}, {\bf 2020-October},
  1585--1589, (2020).

\bibitem{xie2020uda}
Qizhe Xie, Zihang Dai, Eduard Hovy, Thang Luong, and Quoc Le, `Unsupervised
  data augmentation for consistency training', in {\em Advances in Neural
  Information Processing Systems}, eds., H.~Larochelle, M.~Ranzato, R.~Hadsell,
  M.F. Balcan, and H.~Lin, volume~33, pp. 6256--6268. Curran Associates, Inc.,
  (2020).

\bibitem{Yang_Yang_Wang_Xie_2018}
Jun Yang, Runqi Yang, Chongjun Wang, and Junyuan Xie, `Multi-entity
  aspect-based sentiment analysis with context, entity and aspect memory', {\em
  Proceedings of the AAAI Conference on Artificial Intelligence}, {\bf 32}(1),
  (Apr. 2018).

\bibitem{zhuang-2021}
Liu Zhuang, Lin Wayne, Shi Ya, and Zhao Jun, `A robustly optimized {BERT}
  pre-training approach with post-training', in {\em Proceedings of the 20th
  Chinese National Conference on Computational Linguistics}, pp. 1218--1227,
  Huhhot, China, (August 2021). Chinese Information Processing Society of
  China.

\end{thebibliography}
\end{document}